\pdfoutput=1

\documentclass[11pt]{article}

\usepackage[]{acl}

\usepackage{times}
\usepackage{latexsym}

\usepackage[T1]{fontenc}

\usepackage[utf8]{inputenc}

\usepackage{microtype}

\usepackage{inconsolata}

\usepackage{graphicx}

\usepackage{amsfonts}
\usepackage{array}
\usepackage{multirow}
\usepackage{comment}
\usepackage{subcaption}
\usepackage{amsmath}
\usepackage{rotating}
\usepackage[ruled,vlined,linesnumbered,noend]{algorithm2e}
\usepackage{bbding}
\usepackage{soul}

\newcolumntype{B}{>{\centering\arraybackslash}m{3.7cm}}
\newcolumntype{L}{>{\centering\arraybackslash}m{4cm}}
\newcolumntype{X}{>{\centering\arraybackslash}m{6cm}}
\definecolor{darkspringgreen}{rgb}{0.09, 0.45, 0.27}

\title{Know Your Mistakes: Towards Preventing Overreliance on Task-Oriented Conversational AI Through Accountability Modeling}

\author{\textbf{Suvodip Dey, Yi-Jyun Sun, Gokhan Tur, Dilek Hakkani-Tür} \\
  University of Illinois Urbana-Champaign \\
  \texttt{suvodip15@gmail.com, jennysun0830@gmail.com, \{gokhan,dilek\}@illinois.edu}\\
}

\begin{document}
\maketitle
\begin{abstract}
Recent LLMs have enabled significant advancements for conversational agents. However, they are also well known to hallucinate, producing responses that seem plausible but are factually incorrect. On the other hand, users tend to over-rely on LLM-based AI agents, accepting AI’s suggestion even when it is wrong. Adding positive friction, such as explanations or getting user confirmations, has been proposed as a mitigation in AI-supported decision-making systems. In this paper, we propose an accountability model for LLM-based task-oriented dialogue agents to address user overreliance via friction turns in cases of model uncertainty and errors associated with dialogue state tracking (DST). The accountability model is an augmented LLM with an additional accountability head that functions as a binary classifier to predict the relevant slots of the dialogue state mentioned in the conversation. We perform our experiments with multiple backbone LLMs on two established benchmarks (MultiWOZ and Snips). Our empirical findings demonstrate that the proposed approach not only enables reliable estimation of AI agent errors but also guides the decoder in generating more accurate actions. We observe around 3\% absolute improvement in joint goal accuracy (JGA) of DST output by incorporating accountability heads into modern LLMs. Self-correcting the detected errors further increases the JGA from 67.13 to 70.51, achieving state-of-the-art DST performance. Finally, we show that error correction through user confirmations (friction turn) achieves a similar performance gain, highlighting its potential to reduce user overreliance.
\footnote{Code available at \href{https://github.com/uiuc-conversational-ai-lab/Accountable-DST}{github.com/uiuc-conversational-ai-lab/Accountable-DST}}
\end{abstract}

\begin{figure}[t]
    \centering
    \includegraphics[scale=0.7, trim=20 10 20 0, clip]{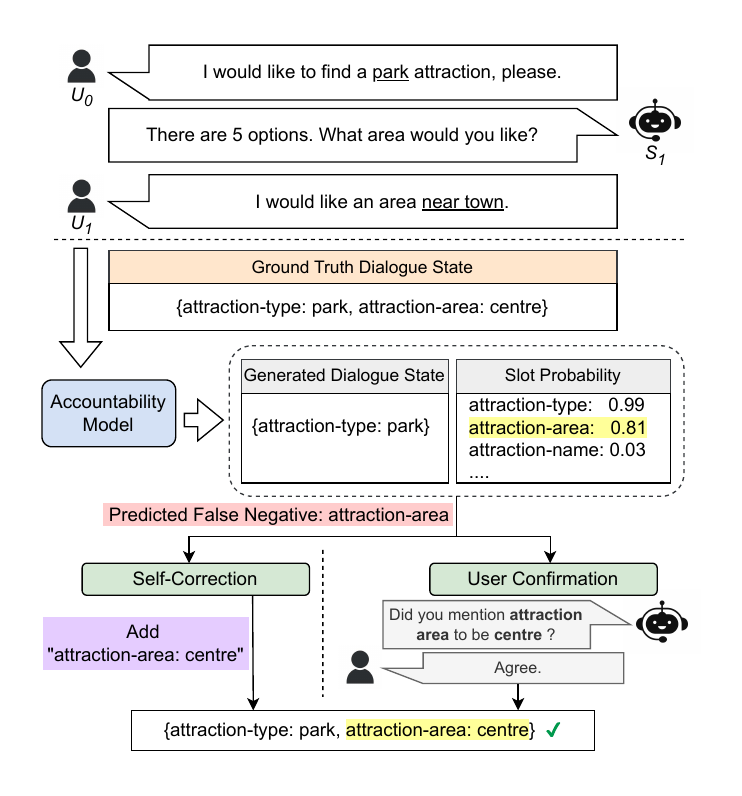}
    \caption{Overview of Accountability Modeling. The model simultaneously generates the dialogue state and estimates the probability of each slot (being included in the dialogue state). Based on these probabilities, it identifies potential errors, which can either be self-corrected using a dedicated algorithm or confirmed by the user through clarification questions.}
    \label{fig:acc_motivation}
    \vspace{-0.2in}
\end{figure}

\section{Introduction}
Conversational agents have reached remarkable advancements with the advent of large language models (LLMs). However, they are prone to hallucinations, generating information that is incorrect or not grounded in reality~\cite{nlg-hal, llm-hal}. At the same time, users often tend to over-rely on LLM-based AI agents, accepting AI suggestions even when they are erroneous~\cite{ai-overreliance, trust-reliance-ai}. In task-oriented conversations, such overreliance causes incorrect or incomplete task execution, undermining the system's reliability. To address this issue, we employ accountability modeling, where accountability refers to the model's ability to explain or justify its actions~\cite{accountability, account-ai}. 
Accountability modeling of task-oriented dialogue systems enables the identification and resolution of errors or unintended consequences, thereby alleviating user overreliance.

Task-oriented dialogue systems (TODS) are designed to assist users in completing a task or goal through conversations. Dialogue state tracking (DST) is a crucial component of TODS that accounts for understanding the user intentions and keeping track of the dialogue state. The dialogue state contains the intents communicated by the user and is generally represented as a set of slot-value pairs. The DST task is to predict the dialogue state after each user turn, as shown in Fig.~\ref{fig:acc_motivation}. 
It can also be seen as a function or API call in end-to-end TODS~\cite{fnctod, autotod}.
There are three types of error associated with DST output:
\begin{itemize}
    \item {\bf False Positives}: Predicted slots that were not mentioned in the dialogue so far.
    \item {\bf False Negatives}: Slots that were mentioned in the dialogue but are missing from the predicted dialogue state.
    \item {\bf Value Errors}: The slot is relevant, but its value is wrong with respect to the dialogue context.
\end{itemize}
Task-oriented dialogues are highly sensitive to these errors, as even a single mistake can significantly alter the conversation's trajectory. LLM-based DST models generally predict the correct slot value for the relevant slots, making value errors a minor concern. However, false positive/negative errors occur more frequently and are critical to task success. For instance, in Fig.~\ref{fig:acc_motivation}, the model initially predicts \texttt{\{attraction-type: park\}} as the dialogue state, causing \texttt{attraction-area} to be a false negative slot for this prediction. As a result, the system may recommend parks that are not near or centre of the town, potentially leading the user to book an unsuitable option due to overreliance on the system. Such issues can degrade the user experience in real-world task-oriented conversations.

In this work, we propose an accountability model for task-oriented conversations to mitigate user overreliance. Our approach aims to enhance the prediction of DST by detecting and correcting false positive and false negative errors, thereby ensuring greater accountability. To achieve this, we integrate an accountability head into the backbone LLMs, which is a binary classifier to predict the slots in the dialogue state. This augmented LLM not only generates the dialogue state but also estimates the probability of each slot being included in the dialogue state. The slot probabilities help to identify the possible false positive/negative slots. These errors are then self-corrected using a dedicated algorithm (Algorithm~\ref{algo1}) that removes false positive slots and adds false negative slots (with the appropriate values), thereby improving the accuracy of DST prediction. For instance in Fig.~\ref{fig:acc_motivation}, 
the ground-truth dialogue state is \texttt{\{attraction-type: park, attraction-area: centre\}}. However, the model initially predicted \texttt{\{attraction-type: park\}}. This results in a false negative, as the prediction does not contain \texttt{attraction-area: centre}. The accountability head assigns a high probability (0.81) to the slot \texttt{attraction-area}, identifying it as a potential false negative. In Fig.~\ref{fig:acc_motivation}, the model self-corrects this error by generating the value for attraction-area (i.e. centre) using Algorithm~\ref{algo1} and adds it to the initial prediction, thereby successfully correcting the dialogue state.

Rather than self-correcting, the conversational agent can also confirm the detected errors with the user through a conversation turn. For example in Fig.~\ref{fig:acc_motivation}, the model asks a clarification question to the user to confirm the error, which helps to rectify the mistake. In recent literature, such mindful interactions or friction turns prompting analytical thinking have been explored to address overreliance on AI~\cite{design-friction, nudge-friction, i̇nan2025betterslowsorryintroducing}. Therefore, slot probabilities from the accountability head can help introduce friction turns, such as confirming model uncertainty and errors, to mitigate user overreliance. The contributions of this work are as follows:
\begin{figure*}[t]
    \centering
    \includegraphics[scale=1, trim=0 5 0 5, clip]{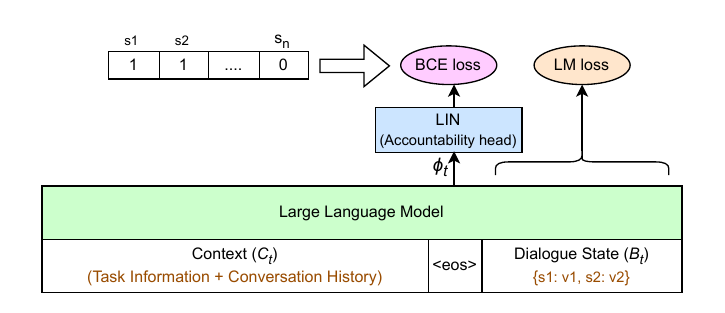}
    \caption{Model architecture of the LLM-based generative accountability modeling for DST.}
    \label{fig:acc_dst}
\vspace{-0.1in}
\end{figure*}
\begin{itemize}
    \item We propose a generative LLM-based accountability model for DST capable of detecting false positives and false negatives.
    \item Accountability modeling improves the DST performance of backbone LLMs on two widely used corpora (MultiWOZ and Snips). For MultiWOZ, the injection of the accountability head in Llama, Mistral, and Gemma shows an absolute improvement of approximately 3\% in joint goal accuracy.
    \item Self-correcting errors identified by the accountability head further enhance the dialogue state prediction, achieving 70.51\% joint goal accuracy and surpassing the performance of state-of-the-art DST models.
    \item Accountability modeling enables the correction of dialogue states via user confirmations which achieves a performance gain comparable to the self-correction strategy, highlighting its potential to reduce user overreliance.
\end{itemize}
\vspace{-0.1in}

\section{Related Work}
DST was initially solved independently from the other TODS modules. In this approach, DST is formulated as a slot-filling task, where the task is to find the relevant slots and then fill out the slot values. Finding the relevant slots is posed as a classification task, while various strategies have been explored to find the slot values like pick-list-based~\cite{nbt, gce, glad}, generative~\cite{trade, som-dst}, and reading comprehension~\cite{gao-reading, trippy, hi-dst}. Although using models with fewer parameters compared to modern LLMs, these methods exhibit decent DST performance. With additional training with synthetic data and sophisticated training, these models have been shown to achieve state-of-the-art DST performance~\cite{assist, metaassist}. However, they are difficult to extend to an end-to-end TODS framework.

With the advancements in Transformer~\cite{transformer} based LLMs, the paradigm shifted towards fully generative and end-to-end modeling of TODS~\cite{simpletod, mintl, dialoglue}. These methods show competitive DST performance compared to the earlier methods. Recently, prompt-based LLM methods have been studied extensively for zero-shot and few-shot DST~\cite{ic-dst, pptod, tods-llm, autotod}. Despite their impressive performance, confidence estimation of dialogue states in generative LLM-based methods is challenging.  LLMs are generally overconfident with the predictions, resulting in high confidence even for spurious predictions~\cite{huang2024large, xiong2024can, thinktwice}. In recent work, \citet{conf-dst} addresses the problem of reliable confidence estimation for DST. However, estimating the confidence for the false negatives is not possible as they are not part of the generated dialogue state~\cite{llm-dst, autotod}. Note that slot-filling-based methods are not susceptible to this issue as they have dedicated classifiers to predict the slots. This work aims to combine the advantages of both approaches for accountability modeling of DST.

\section{Methodology}
This section presents a brief background of LLM-based generative DST, followed by our proposed accountability model and its application in correcting predicted dialogue states.

\subsection{Generative DST}
In this method, DST is solved like standard natural language generation tasks using LLMs. Let $D_t= \{(S_0, U_0), ... (S_n, U_n)\}$ be a task-oriented dialogue where $S_t$ and $U_t$ represent the system and user utterances at turn $t$, respectively. Let $C_t$ represent the dialogue context, which includes $D_t$ as well as optional task-related information and slot descriptions. Let $B_t = \{ s_1: v_1, s_2: v_2, ... \}$ be the dialogue state at turn $t$ where $(s_i, v_i)$ represents the $i^{th}$ slot-value pair. For multi-domain datasets (like MultiWOZ and Snips), $s_i$ is expressed as domain-slot pair. The LLM is trained using the standard language model (LM) loss, defined as follows,  
\begin{equation} \label{eqn:lm}
\mathcal{L}_{\mathrm{LM}} = - \frac{1}{T} \sum_{n=1}^{T} \log p(B_{{t}_{n}}|B_{{t}_{<n}}, C_{t}; \theta)
\end{equation}
where $\theta$ denote the parameters of the LLM and $B_{{t}_{n}}$ is the $n^{th}$ token of the tokenized $B_t$ with $T$ tokens. 

\subsection{Accountability Modeling for DST (AMD)}
In this work, we propose to add an accountability head to make the LLM-based DST generation accountable. The accountability head functions as a classifier to determine whether a slot has been specified in the given context. Let $\phi_{t} \in \mathbb{R}^{d}$ be the encoding of the last token (separator or end-of-sentence) of the context $C_t$, as shown in Fig.~\ref{fig:acc_dst}. Then, the binary-cross entropy (BCE) of the accountability head for a turn $t$ is computed as follows.
\begin{equation} \label{eqn:ffn}
    p = \sigma(\mathrm{LIN}(\phi_t)) \in \mathbb{R}^{|S|}
\end{equation}
\begin{equation} \label{eqn:bce}
    \mathcal{L}_{\mathrm{BCE}} = -\frac{1}{|S|} \sum_{s \in S}^{} (y_s \cdot \log p_s + (1-y_s)\cdot(1-\log p_s))
\end{equation}
where $S$ is the set of slots, $\sigma$ is the element-wise sigmoid operation, $\mathrm{LIN}$ represents a linear layer, $p_s$ denotes the probability of slot $s$, and $y_s \in \{0,1\}$ indicates the label for slot classification denoting the presence/absence of slot $s$ in the ground-truth dialogue state ($B_t$). The final training objective of the accountability model is defined as follows.

\begin{equation} \label{eqn:obj}
    \mathcal{L}_{\mathrm{Account}} = \mathcal{L}_{\mathrm{LM}} + \lambda*\mathcal{L}_{\mathrm{BCE}} 
\end{equation}
where $\lambda \in [0,1]$ is a hyperparameter to control the weight of the BCE loss. Note that we do not generate the dialogue policy or response, as our primary objective is to study the utility of accountability modeling in preventing user overreliance. We plan to explore the complete end-to-end modeling with an accountability head as an extension of this work.


\begin{algorithm}[t]
\DontPrintSemicolon
\SetAlgoLined 
\KwIn{\\$B = $ predicted dialogue state,\\ $S$ = set of slots, 
\\$p$ = slot probabilities, 
\\$\tau_\mathrm{fp} =$ false positive threshold,
\\$\tau_\mathrm{fn} =$ false negative threshold,
\\$D =$ dialogue history.}
\KwOut{Corrected dialogue state ($B'$)}

$B' = \{\}$ \; 
\tcc{\small{Step 1: Filtering false positives}}
\For{$\operatorname{slot}, \operatorname{value} \in B$}{
    \If{$p_{\operatorname{slot}} \geq \tau_\mathrm{fp}$}{
        $B'$[slot] $ \gets \operatorname{value}$\;
    }
}
\tcc{\small{Step 2: Add false negatives}}
$S' \gets$ Set of slots in $B$ \; 
\For{$\operatorname{slot} \in S \setminus S'$}{
    \If{$p_{\operatorname{slot}} \geq \tau_\mathrm{fn}$}{
        $B'$[slot] $\gets$ $\operatorname{generateSlotValue}(D, B,\operatorname{slot})$\;
    }
}
\KwRet $B'$\;
 \caption{Dialogue State Correction}
 \label{algo1}
\end{algorithm}

The accountability head helps estimate the slot probabilities for all the slots, which can be used to detect the false positive and negative slots in the predicted dialogue state. Furthermore, the inclusion of the accountability head acts as an auxiliary loss that helps in the learning of dialogue state generation. Any information about the correct slots can make the dialogue state generation easier. Note that $\phi_t$ is learned to optimize both $\mathcal{L}_{\mathrm{BCE}}$ and $\mathcal{L}_{\mathrm{LM}}$. Therefore, $\phi_t$ encodes the knowledge about the relevant slots, which can improve the accuracy of the generation of the dialogue state. A similar strategy has been shown to be effective in open-domain dialogue generation, where predicting response keywords beforehand can guide and improve response generation~\cite{bok}.

\subsection{Dialogue State Correction using Accountability Model}
\label{sec:bs_correct}
The slot probabilities output by the classifier can be used to self-correct (SC) the generated dialogue state, either by removing slot-value pairs from the dialogue state or forcing the LLM to continue state generation by providing it with the missing slot names. Algo~\ref{algo1} shows the dialogue state correction algorithm. Let $\tau_\mathrm{fp}$ and $\tau_\mathrm{fn}$ be the false positive and false negative thresholds. Let $B$ be the generated dialogue state and $p \in \mathbb{R}^{|S|}$ be the slot probabilities of the classifier. 

The first step of Algo~\ref{algo1} attempts to filter the possible false positives, while the second step helps to include the possible false negatives. Let $\operatorname{generateSlotValue}()$ be the function to generate the slot value for false negative slots (Line 8, Algo~\ref{algo1}). 
The function generates the slot value for a given slot by appending the slot name to the generated dialogue state ($B$) and runs the model decoder to complete the generation. We select the optimal $\tau_\mathrm{fp}$ and $\tau_\mathrm{fn}$ that maximizes the joint goal accuracy of the validation set using grid search.


Instead of self-correcting, we can correct the detected errors through user confirmation. This experimentation is discussed in Section~\ref{sec:overreliance}.

\begin{table}[t]
\begin{small}
\centering
\begin{tabular}{p{1.3cm}|p{0.65cm}|p{1.2cm}|r|r}
\hline \textbf{Dataset} & \textbf{\#Slots} & \textbf{Mode} & \textbf{\#Dialogues} & \textbf{\#Turns}\\ \hline
\multirow{3}{*}{MultiWOZ} & \multirow{3}{*}{30}
& Train & 8420 & 56668\\
&& Validation & 1000 & 7374\\
&& Test & 999 & 7368\\
\hline
\multirow{3}{*}{Snips} & \multirow{3}{*}{53}
& Train & 13084 & 13084\\
&& Validation & 700 & 700\\
&& Test & 700 & 700\\
\hline
\end{tabular}
\caption{\label{tbl:stat1} Data statistics of MultiWOZ and Snips.}
\end{small}
\vspace{-.1in}
\end{table}

\begin{table*}[t]
\begin{small}
\centering
\begin{tabular}{l|p{2.05cm}|lccc|lccc}
\multirow{2}{*}{\textbf{Model}} & \multirow{2}{*}{\textbf{Type}} & \multicolumn{4}{c|}{\textbf{MultiWOZ}}  & \multicolumn{4}{c}{\textbf{Snips}} 
\\ \cline{3-10}
\centering 
& & \textbf{JGA} $\uparrow$ & \textbf{Slot-F1} $\uparrow$  & \textbf{FPR} $\downarrow$ & \textbf{FNR} $\downarrow$ & \textbf{JGA} $\uparrow$ & \textbf{Slot-F1} $\uparrow$  & \textbf{FPR} $\downarrow$ & \textbf{FNR} $\downarrow$ \\ \hline

\multirow{4}{*}{Llama}
& $\mathcal{M_{\operatorname{SFT}}}$ & 64.34 & 95.23 & 12.17 & 23.72 &  92.43 & 97.76 & 5.14 & 4.71 \\ 
& $\mathcal{M_{\operatorname{AMD}}}$ & 67.13 \textcolor{blue}{\textsubscript{$\uparrow$  4.3}} &  95.90 & 13.17 & 18.28 & 93.57 \textcolor{blue}{\textsubscript{$\uparrow$ 1.2}} & 98.00 & 4.71 & 4.43 \\ 
& $\mathcal{M_{\operatorname{AMD+SC}}}$ & {\bf 70.51} \textcolor{blue}{\textsubscript{$\uparrow$  9.6}} & {\bf 96.51} & 12.83 & 14.44 & 93.71 \textcolor{blue}{\textsubscript{$\uparrow$ 1.4}} & 98.17 & 4.00 & 4.00\\ 
\hline
\multirow{4}{*}{Mistral}
& $\mathcal{M_{\operatorname{SFT}}}$ & 65.86  & 95.68 & 11.90 & 20.41 & 92.57 & 97.57 & 6.28 & 5.28 \\ 
& $\mathcal{M_{\operatorname{AMD}}}$ & 68.58 \textcolor{blue}{\textsubscript{$\uparrow$  4.1}} & 96.19 & 11.35 & 16.94 & 93.71 \textcolor{blue}{\textsubscript{$\uparrow$  1.2}} & 98.06 & 4.85 & 4.43 \\ 
& $\mathcal{M_{\operatorname{AMD+SC}}}$ & 69.84 \textcolor{blue}{\textsubscript{$\uparrow$  6.0}} & 96.37 & 12.74 & 14.19 & {\bf 94.00} \textcolor{blue}{\textsubscript{$\uparrow$  1.5}} & {\bf 98.17} & 4.57 & 4.14 \\ 
\hline
\multirow{4}{*}{Gemma}
& $\mathcal{M_{\operatorname{SFT}}}$ & 62.12 & 95.05 & 6.35 & 28.84 & 91.43 & 97.37 & 5.71 & 5.29 \\ 
& $\mathcal{M_{\operatorname{AMD}}}$ & 65.05 \textcolor{blue}{\textsubscript{$\uparrow$  4.7}} &  95.68 & 12.47 & 20.15  & 91.86 \textcolor{blue}{\textsubscript{$\uparrow$  0.5}} & 97.86 & 5.43 & 5.29 \\ 
& $\mathcal{M_{\operatorname{AMD+SC}}}$ & 66.27 \textcolor{blue}{\textsubscript{$\uparrow$  6.7}} & 96.03 & 16.08 & 15.08 & 92.00 \textcolor{blue}{\textsubscript{$\uparrow$  0.6}} & 98.00 & 5.14 & 5.14 \\ 
\hline

\end{tabular}
\caption{Comparison of the DST performance on the MultiWOZ 2.4 and Snips test datasets with different LLM backbones. The relative JGA improvement of proposed $\mathcal{M_{\operatorname{AMD}}}$ and $\mathcal{M_{\operatorname{AMD+SC}}}$, compared to the respective $\mathcal{M_{\operatorname{SFT}}}$ baseline, is highlighted in blue. The best results are indicated in bold font.}
\label{tbl:res1}
\end{small}
\end{table*} 

\section{Experiment Setup}
\subsection{Dataset}
We experiment with two widely used datasets: i) MultiWOZ~\cite{multiwoz}, and ii) Snips~\cite{snips}. MultiWOZ is one of the largest multi-domain conversation corpora for task-oriented dialogue, containing multi-turn conversations. We use MultiWOZ 2.4~\cite{multiwoz24}, which contains fewer annotation errors and inconsistencies than the other MultiWOZ versions. On the other hand, Snips is a similar dataset for spoken language understanding, containing only single-turn conversations. The basic statistics of both datasets are shown in Table~\ref{tbl:stat1}.

\subsection{Evaluation Metric}

DST is primarily evaluated using joint goal accuracy (JGA). JGA is defined as the percentage of turns where the predicted dialogue state exactly matches the ground-truth~\cite{dstc2, fga}. Additionally, we report the Slot-F1 score, which measures the slot-level F1 performance. We also analyze false positive rate (FPR) and false negative rate (FNR) for certain results. FPR and FNR are defined as the percentage of turns containing false positive and false negative slots, respectively. Let $X$ be the number of turns containing at least one false positive slot. Let $Y$ be the number of turns containing at least one false negative slot. Then, FPR and FNR are defined as $\operatorname{FPR}=\frac{100X}{T}$ and $\operatorname{FNR}=\frac{100Y}{T}$, where $T$ is the total number of turns in the dataset.

\subsection{Model Architectures and Variants}
\label{sec:variant}
We study the utility of the proposed accountability model by applying it to three LLMs - Llama 3.1 (8B)~\cite{llama}, Mistral (7B)~\cite{mistral}, and Gemma (7B)~\cite{gemma}. We use the instruction-tuned version of the three models. The model nomenclature used in our experiments is described as follows.
\begin{itemize}
    \item $\mathcal{M_{\operatorname{SFT}}}$: Model trained using supervised fine-tuning (SFT) for dialogue state generation with only $\mathcal{L_{\mathrm{LM}}}$ (Eqn.~\ref{eqn:lm}), which is methodologically similar to LDST~\cite{ldst} with minor modifications.\footnote{LDST~\cite{ldst} is trained to generate the value for a single slot where all possible slot-values from the database is provided in the context as meta-data, limiting its scalability. In contrast, $\mathcal{M_{\operatorname{SFT}}}$ is trained to generate the full dialogue state without using any such meta-data.}
    \item $\mathcal{M_{\operatorname{AMD}}}$: Proposed \textbf{A}ccountability \textbf{M}odel for \textbf{D}ST (AMD) with accountability head, fine-tuned with $\mathcal{L_{\mathrm{Account}}}$ (Eqn.~\ref{eqn:obj}). 
    \item $\mathcal{M_{\operatorname{AMD+SC}}}$: $\mathcal{M_{\operatorname{AMD}}}$ after \textbf{S}elf-\textbf{C}orrecting (SC) the dialogue state using Algo~\ref{algo1}.
\end{itemize}

\subsection{Training Details}
\label{sec:exp}
All the models are implemented using the PyTorch and Huggingface libraries in Python 3.12.
We used LoRA~\cite{lora} finetuning with rank ($r$) $8$, $\alpha=32$, and dropout $0.1$. We used AdamW optimizer with learning rate 5e-5 to fine-tune both $\mathcal{M_{\operatorname{SFT}}}$ and $\mathcal{M_{\operatorname{AMD}}}$. We trained all the models for 4 epochs and chose the final model with the minimum validation loss. The prompts used for model finetuning are provided in Appendix~\ref{sec:prompt}. In Eqn.~\ref{eqn:obj}, the optimal $\lambda$ is selected based on the JGA score of the validation set. The best $\lambda$ was found to be 0.25 except for Mistral-Snips and Gemma-Snips, where $\lambda=0.1$ resulted in the best validation performance. For $\mathcal{M_{\operatorname{AMD+SC}}}$, the false positive ($\tau_{\operatorname{fp}}$) and the false negative ($\tau_{\operatorname{fn}}$) threshold is selected similarly based on the validation performance. We observed that (0.1, 0.5) and (0.05, 0.9) are the best ($\tau_{\operatorname{fp}}$, $\tau_{\operatorname{fn}}$) for MultiWOZ and Snips, respectively. These optimal values of $\lambda$, $\tau_{\operatorname{fp}}$, and $\tau_{\operatorname{fn}}$ are used to show the test performance in the rest of the article.




Further details on prompts, training, and inference are provided in Section~\ref{sec:prompt}.

\section{Results and Analysis}

\subsection{DST Performance}
Table~\ref{tbl:res1} compares the DST performance of the different model variants described in Section~\ref{sec:variant}. Table~\ref{tbl:res_old_baselines} shows the comparison with the previous baselines. 
The observations are summarized as follows.

\textbf{Impact of Accountability Head}: In Table~\ref{tbl:res1}, we observe that adding the accountability head in $\mathcal{M_{\operatorname{AMD}}}$ improves the performance of $\mathcal{M_{\operatorname{SFT}}}$. Both joint goal accuracy (JGA) and Slot-F1 are improved for all the backbone models in both datasets. Overall, in MultiWOZ, we can observe an absolute improvement of around 3\%. Similarly, we observe approximately 1\% improvement for Snips. The margin of improvement in Snips is lower as it is relatively easier than MultiWOZ. 
The improvement in JGA with the accountability head can be attributed to the significant reduction in FNR for MultiWOZ and the reduction in both FPR and FNR for Snips.

\begin{table}[t]
\begin{small}
\centering
\begin{tabular}{p{1.1cm} | p{4.8cm} | r}
\textbf{Type} & \textbf{Model} & \textbf{JGA} \\ \hline
\multirow{1}{*}{Zero-Shot}
& GPT-4o~\cite{conf-dst} & 36.10 \\
\hline
\multirow{2}{*}{Few-Shot}
& IC-DST~\cite{ic-dst} & 62.43 \\ 
& OrchestraLLM~\cite{orchestrallm} & 52.68 \\
& CorrectionLM~\cite{correctionlm} & 57.35 \\
\hline
\multirow{9}{*}{SFT}
& TRADE~\cite{trade} & 55.05 \\
& SUMBT~\cite{sumbt} & 61.86 \\
& SimpleTOD~\cite{simpletod} & 57.18 \\
& TripPy~\cite{trippy} & 64.75 \\
& SOM-DST~\cite{som-dst} & 66.78 \\
& Seq2Seq~\cite{zhao-etal-2021-effective-sequence} & 67.10 \\
& TripPy-R~\cite{heck-etal-2022-robust} & 69.87 \\
& LDST~\cite{ldst} (10\% data) &  62.45 \\
& $\mathcal{M_{\operatorname{AMD}}}$ (Ours) & 67.13 \\
& \textbf{$\pmb{\mathcal{M_{\operatorname{AMD+SC}}}}$ (Ours)} & \textbf{70.51} \\
\hline

\end{tabular}
\caption{JGA comparison between various DST models on the MultiWOZ 2.4 test data.}
\label{tbl:res_old_baselines}
\end{small}
\vspace{-0.1in}
\end{table} 

\textbf{Impact of Self-Correction}: $\mathcal{M_{\operatorname{AMD+SC}}}$ self-corrects the predicted dialogue states of $\mathcal{M_{\operatorname{AMD}}}$ using Algo~\ref{algo1}. We use the optimal $\tau_{\operatorname{fp}}$ and $\tau_{\operatorname{fn}}$ thresholds that maximize the JGA of the validation set (as described in Section~\ref{sec:exp}). We can observe that self-correcting the dialogue state improves DST performance significantly. Considering the best results in Table~\ref{tbl:res1}, $\mathcal{M_{\operatorname{AMD+SC}}}$ improves the performance of the baseline $\mathcal{M_{\operatorname{SFT}}}$  model by 9.6\% and 1.5\% for MultiWOZ and Snips, respectively. We also observe that correcting false negatives significantly reduces the FNR in both MultiWOZ and Snips. However, false negative correction can sometimes introduce spurious slots (discussed in Section~\ref{sec:ab1}), leading to a higher FPR in $\mathcal{M_{\operatorname{AMD+SC}}}$ compared to $\mathcal{M_{\operatorname{AMD}}}$ in certain MultiWOZ instances where false negatives are more prevalent than false positives. 
This happens because we optimized $\tau_{\operatorname{fp}}$ and $\tau_{\operatorname{fn}}$  to maximize JGA rather than FPR or FNR. In Snips, self-correction reduces both FPR and FNR because of the more balanced occurrence of false positives and false negatives.

\textbf{Comparison with previous baselines}:
Table~\ref{tbl:res_old_baselines} compares our proposed $\mathcal{M_{\operatorname{AMD}}}$ and $\mathcal{M_{\operatorname{AMD+SC}}}$ with various zero-shot, few-shot, and supervised fine-tuning (SFT) baselines on the MultiWOZ 2.4 dataset. The results demonstrate that $\mathcal{M_{\operatorname{AMD+SC}}}$ outperforms both zero-shot and few-shot baselines and achieves state-of-the-art performance compared to SFT baselines. For fair comparison, we do not consider models such as STAR~\cite{star} and ASSIST~\cite{assist} that use additional synthetic data for training and generally achieve higher JGA (around 80\%). 

\subsection{Impact of Varying False Positive and False Negative Threshold}
\label{sec:ab1}
In this section, we study the impact of varying the false positive threshold ($\tau_{\operatorname{fp}}$) and false negative threshold ($\tau_{\operatorname{fn}}$) while correcting the dialogue states using Algo~\ref{algo1}. Table~\ref{tbl:fpfn} shows the change in DST performance by varying $\tau_{\operatorname{fp}}$ and $\tau_{\operatorname{fn}}$. In our setup,  $\tau_{\operatorname{fp}}=0$ and $\tau_{\operatorname{fn}}=1$ indicate no correction for false positives and false negatives, respectively. 

\begin{table}[t]
\begin{small}
\centering
\begin{tabular}{p{1.1cm}|p{0.35cm}|p{0.3cm}|p{0.4cm}|rrr}
\textbf{Type} & \textbf{$\tau_{\operatorname{fp}}$} & \textbf{$\tau_{\operatorname{fn}}$} & \textbf{Cost} & \textbf{JGA} $\uparrow$  & \textbf{FPR} $\downarrow$ & \textbf{FNR} $\downarrow$\\ \hline
$\mathcal{M_{\operatorname{AMD}}}$ & 0 & 1 & 0 & 67.13 & 13.17 & 18.28 \\ 
\hline
\multirow{3}{1.3cm}{Filter False Positives}
& 0.1 & 1 & 0 & 68.15 & 11.16 & 18.92 \\ 
& 0.2 & 1 & 0 & 68.01 & 9.54 & 20.63 \\ 
& 0.3 & 1 & 0 & 67.24 & 8.24 & 22.86 \\
\hline
\multirow{4}{1.3cm}{Add False Negatives}
& 0 & 0.9 & 1.5 & 67.55 & 13.59 & 17.19 \\ 
& 0 & 0.7 & 4.7 & 68.59 & 14.68 & 14.60 \\ 
& 0 & 0.5 & 7.5 & 69.31 & 16.39 & 12.11 \\
& 0 & 0.4 & 8.9 & 68.97 & 18.28 & 10.74 \\
\hline

\end{tabular}
\caption{Impact of false positive and negative threshold on dialogue state correction in MultiWOZ 2.4 test set with Llama. ``Cost'' denotes the \%turns where the model generated values for false negative correction.}
\label{tbl:fpfn}
\end{small}
\vspace{-0.1in}
\end{table}

Firstly, we study the impact of filtering only possible false positive slots whose probabilities are less than $\tau_{\operatorname{fp}}$ (Line 3-4, Algo~\ref{algo1}). We can observe that increasing $\tau_{\operatorname{fp}}$ reduces FPR. However, this process also filters correct slots with $p_{\operatorname{slot}} < \tau_{\operatorname{fp}}$, thereby increasing the FNR. As a result, we can observe that the JGA starts degrading when $\tau_{\operatorname{fp}}>0.1$ due to increasing FNR. Since we are discarding slots based on a threshold, no additional cost is involved in generating slot values.

Next, we analyze the impact of correcting only possible false negative slots with different $\tau_{\operatorname{fn}}$ (Line 5-8, Algo~\ref{algo1}). We can observe that decreasing $\tau_{\operatorname{fn}}$ reduces FNR. However, this process also tries to include false positive slots with $p_{\operatorname{slot}} \geq \tau_{\operatorname{fn}}$, thereby increasing the FPR. Consequently, JGA drops when $\tau_{\operatorname{fn}}<0.5$. 
Note that correcting false negatives involves generating the slot values. In Table~\ref{tbl:fpfn}, we denote this generation cost by the percentage of turns that require slot-value generation to rectify false negatives. This generation cost is inversely proportional to $\tau_{\operatorname{fn}}$. On average, slot values are generated for 1.1 slots per updated turn.

\begin{figure}[t]
    \centering
    \includegraphics[scale=0.50]{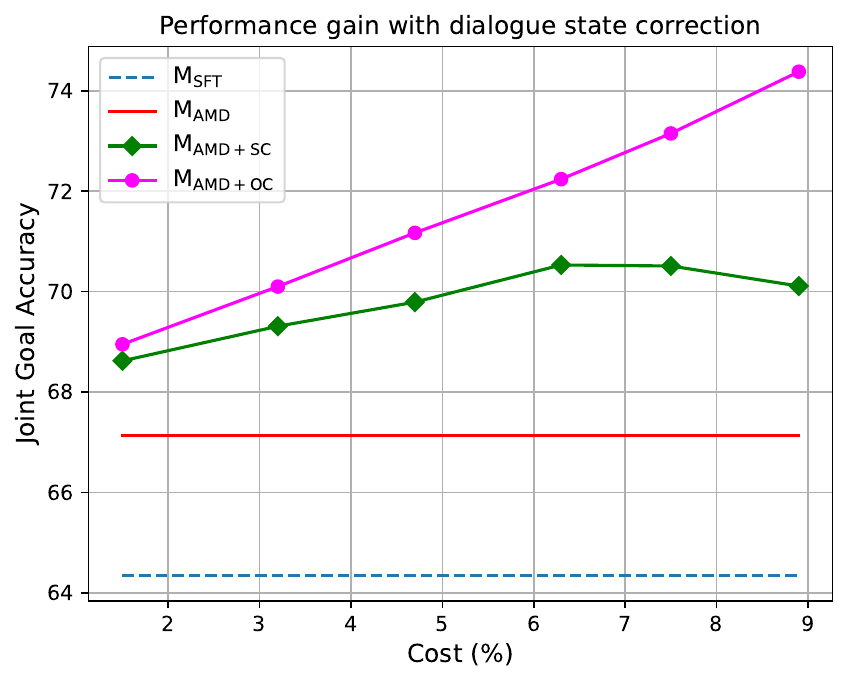}
    \caption{Performance gain (JGA) with varying $\tau_{\operatorname{fp}}$ and $\tau_{\operatorname{fn}}$ on MultiWOZ 2.4 test set with Llama. The $x$-axis shows the \%turns with false negative correction. On average, 1.1 slots have been involved in the false negative corrections consistently over the range of this plot. Hence, we did not include it in the depiction.}
    \label{fig:corrected_jga}
\vspace{-0.1in}
\end{figure}

Fig.~\ref{fig:corrected_jga} shows the summary of the performance gain achieved with dialogue state correction. We observe that $\mathcal{M_{\operatorname{AMD+SC}}}$ significantly enhances the performance of $\mathcal{M_{\operatorname{SFT}}}$ (9.6\% relative) through self-corrections. The best performance in MultiWOZ is achieved with $\tau_{\operatorname{fp}}=0.1$ and $\tau_{\operatorname{fn}}=0.5$. 
However, beyond a certain point, the JGA begins to decline as false negative corrections induce false positives, as previously discussed.  Figure~\ref{fig:corrected_jga} also depicts the maximum JGA achievable through dialogue state correction using $\mathcal{M_{\operatorname{AMD+OC}}}$ where OC stands for \textbf{O}racle-\textbf{C}orrection. Instead of self-correcting, $\mathcal{M_{\operatorname{AMD+OC}}}$ corrects the output of $\mathcal{M_{\operatorname{AMD}}}$ using the oracle or ground-truth slot values in Step 2 of Algo~\ref{algo1}. $\mathcal{M_{\operatorname{AMD+OC}}}$  does not introduce false positives while addressing false negatives, as only the slots that correspond to the ground truth are updated. The plot suggests that there is still room to achieve higher JGA through self-correction.

\begin{table}[t]
\begin{small}
\centering
\begin{tabular}{r|rr|rr}
\multirow{2}{*}{\textbf{$\lambda$}} & \multicolumn{2}{c|}{\textbf{MultiWOZ}}  & \multicolumn{2}{c}{\textbf{Snips}} 
\\ \cline{2-5}
\centering 
 & \textbf{JGA} & \textbf{Slot-F1} & \textbf{JGA} & \textbf{Slot-F1} \\ \hline
0 & 64.34 & 95.23 & 92.43 & 97.76 \\ 
0.1 & 65.93 & 95.64 & 92.71 & 97.82 \\ 
0.25 & 67.13 & 95.90 & 93.57 & 98.01 \\ 
0.5 & 66.27 &  95.80 & 93.14 & 97.87  \\ 
0.75 & 65.40  & 95.61 & 93.00 & 97.84 \\ 
1.0 & 65.25 & 95.57 & 92.86 & 97.72  \\ 
\hline

\end{tabular}
\caption{Impact of varying $\lambda$ in $\mathcal{M_{\operatorname{AMD}}}$ with Llama backbone on test data.}
\label{tbl:ab1}
\end{small}
\vspace{-0.1in}
\end{table} 

\subsection{Impact $\lambda$ on DST performance}
\label{sec:ab2}
In this ablation study, we vary the value of $\lambda$ in Eqn.~\ref{eqn:obj}. In Table~\ref{tbl:ab1}, for both datasets, we observe that the JGA and Slot-F1 increase with $\lambda$ initially and then start dropping. Note that the primary task here is to generate the dialogue state for which the JGA or Slot-F1 is estimated. Now, increasing the $\lambda$ increases the importance of the accountability head, which starts affecting the generation head after a certain point. Therefore, although any $\lambda \in [0,1]$ can improve the base model, a thorough hyperparameter optimization of $\lambda$ is required to achieve the best possible DST performance. The validation performance used to select the optimal value of $\lambda$ is discussed in Appendix~\ref{sec:hyp}.

\begin{table*}[t]
\begin{small}
\centering
\begin{tabular}{m{1.5cm}|m{4.8cm}|m{3.5cm}|m{3cm}|m{1cm}}
\textbf{ID} & \textbf{Dialogue History} & \textbf{Generated dialogue state} & \textbf{Predicted Errors} & \textbf{Post-Correct Status} \\ \hline

\multirow{3}{1.5cm}{Example \#1 PMUL3514 (MultiWOZ)} & $U_0$: I am looking for information in Cambridge. 
& \multirow{3}{3.5cm}{\{restaurant-pricerange: expensive, hotel-pricerange: moderate, hotel-stars: 0\} 
\textcolor{red}{\XSolidBrush}}  & \multirow{3}{3cm}{False Positive:- restaurant-pricerange: expensive (0.02) } & \multirow{3}{1.2cm}{\textcolor{darkspringgreen}{\Checkmark}} \\
\centering
& $S_1$: I need more specifics to help you. what type of information do you need? & & & \\
& $U_1$: I would like a moderately priced place to stay. but only if it is a 0 star. I love a little adventure! & & & \\
\hline


\multirow{3}{1.5cm}{Example \#2 PMUL3919 (MultiWOZ)} & $U_0$: Can you find a theater to go to in town? 
& \multirow{3}{3.5cm}{\{attraction-type: theatre\} \textcolor{red}{\XSolidBrush}} & \multirow{3}{3cm}{False Negative:- attraction-area: dontcare (0.6) } & \multirow{3}{1.5cm}{\textcolor{darkspringgreen}{\Checkmark}} \\
\centering
& $S_1$: Sure, do you have an area of town you would like to visit? & & & \\
& $U_1$: No, I am not concerned about that. & & & \\
\hline

\multirow{5}{1.5cm}{Example \#3 MUL1289 (MultiWOZ)} & $U_0$: I am trying to plan a trip there but need a cheap place to stay. 
& \multirow{3}{3.5cm}{\{hotel-area: west, hotel-pricerange: cheap, hotel-internet: yes, hotel-type: hotel\} \textcolor{red}{\XSolidBrush}} & \multirow{3}{3cm}{False Positive:- hotel-type: hotel (0.06)  \hspace{1cm}    False Negative:- hotel-name: finches bed and breakfast (0.72) } & \multirow{3}{1.5cm}{\textcolor{darkspringgreen}{\Checkmark}} \\
\centering
& $S_1$: Alexander bed and breakfast is located in the centre. They are located at 56 saint barnabas road. & & & \\
& $U_1$: Hmm, i am looking for a place in the west. It does not need to include internet. & & & \\
& $S_2$: finches bed and breakfast is cheap. & & & \\
& $U_2$: okay do they have free wifi? & & & \\
\hline

\multirow{3}{1.5cm}{Example \#4 PMUL1091 (MultiWOZ)} & $U_0$: Can you locate for me a train that leaves on tuesday after 3:15 pm? thanks. 
& \multirow{3}{3.5cm}{\{train-leaveat: 15:15, train-day: tuesday\} \textcolor{darkspringgreen}{\Checkmark}} & \multirow{3}{3cm}{False Negative:- train-departure: cambridge (0.75) } & \multirow{3}{1.5cm}{\textcolor{red}{\XSolidBrush}} \\
\centering
& $S_1$: There is a train that leaves cambridge at 15:00 and arrives at london kings cross at 15:51 on tuesday. & & & \\
& $U_1$: That's too early. I need to leave after 15:15. & & & \\
\hline

\multirow{3}{1.5cm}{Example \#5 MUL2053 (MultiWOZ)} & $U_0$: Hi there. Can you help me find a 2 star rated hotel or guest house? 
& \multirow{3}{3.5cm}{\{hotel-stars: 2, hotel-internet: yes\} \textcolor{red}{\XSolidBrush}} & \multirow{3}{3cm}{False Negative:- hotel-name: ashley hotel (0.81), hotel-type: hotel (0.52) } & \multirow{3}{1.5cm}{\textcolor{red}{\XSolidBrush}} \\
\centering
& $S_1$: Ashley hotel is a 2 star hotel in 74 chesterton road. & & & \\
& $U_1$: Does that include wifi? & & & \\
\hline

\end{tabular}
\caption{Illustrative example of dialogue state corrections using the proposed $\mathcal{M_{\operatorname{AMD+SC}}}$ model.}
\label{tbl:example}
\end{small}
\vspace{-0.1in}
\end{table*} 

\subsection{Qualitative Analysis}
In this section, we provide a qualitative analysis of the $\mathcal{M_{\operatorname{AMD+SC}}}$ model. Table~\ref{tbl:example} shows illustrative examples of the model prediction from the MultiWOZ datasets. In the first example, the model detected a false positive slot (\textit{restaurant-pricerange}) and filtered it to rectify the prediction. The second example contains a false negative slot (\textit{attraction-type}), which is corrected successfully. In the third example, the model detects both false positive and false negative slots and successfully rectifies them. In the fourth example, the original prediction of the model is correct. However, self-correction adds an extra slot (\textit{train-departure}), which makes the prediction wrong. The fifth one shows an instance where the algorithm partially corrects an error. Here, the model included \textit{hotel-name} to its prediction but added a false positive (\textit{hotel-type}) in the process because $\tau_{\operatorname{fn}}$ was set to 0.5. This example demonstrates how addressing false negatives can sometimes introduce false positives.

\section{Preventing User Overeliance with Accountability Model}
\label{sec:overreliance}
So far, we have focused on analyzing the capability of the proposed accountability model in detecting errors and self-correcting them to improve DST performance. In this section, we discuss its application to prevent user overreliance on task-oriented conversational AI. The approach is based on introducing positive friction~\cite{i̇nan2025betterslowsorryintroducing} like user confirmations that can eventually lead to successful task completion. Let $E_t $ be the set of erroneous slot-value pairs predicted by $\mathcal{M_{\operatorname{AMD}}}$. Given $E_t$, a task-oriented conversational agent can introduce friction turns to get clarification on the unconfident slots. Introducing such friction turns have been shown to be helpful in preventing user overreliance on AI~\cite{nudge-friction, i̇nan2025betterslowsorryintroducing}. 

To evaluate this method, we conduct an experiment using a user simulator. We employ ``GPT-4o mini''~\cite{openai2024gpt4technicalreport} as our user simulator, assuming cooperative user behavior. Given the dialogue history and a slot-value pair (detected as false negative or false positive), the user simulator is asked to confirm if it mentioned the slot-value pair by responding ``Agree'' (confirms interest) or ``Disagree'' (rejects). 
We correct the dialogue state by removing the false positive slots and adding the false negative slots confirmed by the user. A detailed description of the experimental setup is provided in Appendix~\ref{sec:prompt}. 


Table~\ref{tbl:usersim} shows the results of this experiment where $\mathcal{M_{\operatorname{AMD+UC}}}$ denotes the \textbf{U}ser \textbf{C}onfirmation (UC) provided by the simulator. We can observe that $\mathcal{M_{\operatorname{AMD+UC}}}$ achieves comparable performance to $\mathcal{M_{\operatorname{AMD+SC}}}$. 
Although the user simulator may not fully reflect actual user behavior, assuming a cooperative user allows us to interpret the results in Table~\ref{tbl:usersim} as a lower bound on achievable DST performance. A real cooperative user would likely be even more effective in identifying and confirming errors, potentially leading to further performance improvements. Additional improvements may also be possible through the use of more advanced simulators and refined prompt engineering. Nonetheless, this experiment demonstrates the practical value of our accountability modeling framework in reducing user overreliance in real-world task-oriented dialogues.

\begin{table}[t]
\begin{small}
\centering
\begin{tabular}{l|l|rrrrr|rrrrr}
\textbf{Model} & \textbf{Type} & \textbf{JGA} & \textbf{Slot-F1}\\ \hline

\multirow{3}{*}{Llama}
& $\mathcal{M_{\operatorname{AMD}}}$ & 67.13 & 95.90 \\ 
& $\mathcal{M_{\operatorname{AMD+SC}}}$ & 70.49 & 96.51 \\ 
& $\mathcal{M_{\operatorname{AMD+UC}}}$ & \textbf{70.78} & \textbf{96.55} \\ 
\hline
\multirow{3}{*}{Mistral}
& $\mathcal{M_{\operatorname{AMD}}}$ & 68.58  & 96.19 \\ 
& $\mathcal{M_{\operatorname{AMD+SC}}}$ & 69.84 & 96.37 \\ 
& $\mathcal{M_{\operatorname{AMD+UC}}}$ & 70.21 & 96.41 \\ 
\hline
\multirow{3}{*}{Gemma}
& $\mathcal{M_{\operatorname{AMD}}}$ & 65.05 & 95.68 \\ 
& $\mathcal{M_{\operatorname{AMD+SC}}}$ & 66.27 & 96.03 \\ 
& $\mathcal{M_{\operatorname{AMD+UC}}}$ & 66.32 & 96.05 \\ 
\hline

\end{tabular}
\caption{Impact of correcting dialogue state using user simulator on MultiWOZ test data. SC and UC denote self-correction and user confirmation, respectively.}
\label{tbl:usersim}
\end{small}
\vspace{-0.1in}
\end{table} 

\section{Conclusion}
In conclusion, we present an LLM-based generative accountability modeling for task-oriented dialogue systems. The core idea of our approach involves incorporating an accountability head into backbone LLMs, which functions as a binary classifier to predict the slots in the dialogue state. Doing so not only enables the detection of both false positives and false negatives but also guides the generation of accurate dialogue states. We empirically show that accountability modeling improves the DST performance of backbone LLMs (Llama, Mistral, and Gemma) on two widely used task-oriented corpora (MultiWOZ and Snips). 
Identifying the errors also enables self-correction of the dialogue state, which helps to achieve state-of-the-art performance. 
Finally, we demonstrate the utility of the proposed accountability modeling to correct the DST errors in an interactive setup via user confirmations, thereby preventing user overreliance. In the future, we want to extend accountability modeling for end-to-end task-oriented conversations. 



\section{Limitations}
We recognize the following limitations regarding our work.
\begin{itemize}
    \item The accountability model assumes the data to be annotated along with a fixed number of domains and slots. This is why the proposed model cannot be directly used for new or unseen domains/slots. However, in task-oriented conversations, domains and slots are typically well-defined, making the model effectively trainable. Furthermore, the accountability model outperforms few-shot and zero-shot methods, as shown in Table~\ref{tbl:res_old_baselines}. While general LLMs (such as GPT-4o) can operate with limited data, their performance remains suboptimal. In task-oriented conversations, accurately understanding intent is essential for effective dialogue management. Moreover, disregarding the training overhead, a small LLM with accountability modeling is more cost-effective than a large general-purpose LLM for dialogue state tracking in real applications.
    \item In this work, the notion of accountability is added by estimating the errors in the DST task. Hence, the proposed accountability modeling approach is specific to sequence tagging tasks like DST, information extraction, entity extraction, etc.

    
\end{itemize}

\section{Ethics Statement}
This work proposes accountability modeling for task-oriented conversational agents. We use the publicly available MultiWOZ and Snips datasets in full compliance with their terms of use. Our experiments do not use any private, confidential, or real personal data. Our model's use of personal information is limited to the task-oriented conversation of MultiWOZ and Snips. Since we use LLMs to generate dialogue states, there is minimal risk of generating harmful, biased, or discriminatory statements. We acknowledge such potential ethical concerns associated with this work.

\section{Acknowledgments}

This work was supported in part by Other Transaction award HR0011249XXX from the U.S. Defense Advanced Research Projects Agency (DARPA) Friction for Accountability in Conversational Transactions (FACT) program and has benefited from the Microsoft Accelerate Foundation Models Research (AFMR) grant program, through which leading foundation models hosted by Microsoft Azure and access to Azure credits were provided to conduct the research.

\bibliography{main}

\begin{thebibliography}{51}
\providecommand{\natexlab}[1]{#1}

\bibitem[{Budzianowski et~al.(2018)Budzianowski, Wen, Tseng, Casanueva, Ultes, Ramadan, and Ga{\v{s}}i{\'c}}]{multiwoz}
Pawe{\l} Budzianowski, Tsung-Hsien Wen, Bo-Hsiang Tseng, I{\~n}igo Casanueva, Stefan Ultes, Osman Ramadan, and Milica Ga{\v{s}}i{\'c}. 2018.
\newblock \href {https://doi.org/10.18653/v1/D18-1547} {{M}ulti{WOZ} - a large-scale multi-domain wizard-of-{O}z dataset for task-oriented dialogue modelling}.
\newblock In \emph{Proceedings of the 2018 Conference on Empirical Methods in Natural Language Processing}, pages 5016--5026, Brussels, Belgium. Association for Computational Linguistics.

\bibitem[{Coucke et~al.(2018)Coucke, Saade, Ball, Bluche, Caulier, Leroy, Doumouro, Gisselbrecht, Caltagirone, Lavril, Primet, and Dureau}]{snips}
Alice Coucke, Alaa Saade, Adrien Ball, Th{\'e}odore Bluche, Alexandre Caulier, David Leroy, Cl{\'e}ment Doumouro, Thibault Gisselbrecht, Francesco Caltagirone, Thibaut Lavril, Ma{\"e}l Primet, and Joseph Dureau. 2018.
\newblock \href {https://api.semanticscholar.org/CorpusID:44061213} {Snips voice platform: an embedded spoken language understanding system for private-by-design voice interfaces}.
\newblock \emph{ArXiv}, abs/1805.10190.

\bibitem[{Dey and Desarkar(2021)}]{hi-dst}
Suvodip Dey and Maunendra~Sankar Desarkar. 2021.
\newblock \href {https://aclanthology.org/2021.sigdial-1.23} {Hi-{DST}: A hierarchical approach for scalable and extensible dialogue state tracking}.
\newblock In \emph{Proceedings of the 22nd Annual Meeting of the Special Interest Group on Discourse and Dialogue}, pages 218--227, Singapore and Online. Association for Computational Linguistics.

\bibitem[{Dey and Desarkar(2024)}]{bok}
Suvodip Dey and Maunendra~Sankar Desarkar. 2024.
\newblock \href {https://doi.org/10.18653/v1/2024.sigdial-1.48} {{B}o{K}: Introducing bag-of-keywords loss for interpretable dialogue response generation}.
\newblock In \emph{Proceedings of the 25th Annual Meeting of the Special Interest Group on Discourse and Dialogue}, pages 566--578, Kyoto, Japan. Association for Computational Linguistics.

\bibitem[{Dey et~al.(2022)Dey, Kummara, and Desarkar}]{fga}
Suvodip Dey, Ramamohan Kummara, and Maunendra Desarkar. 2022.
\newblock \href {https://doi.org/10.18653/v1/2022.acl-short.35} {Towards fair evaluation of dialogue state tracking by flexible incorporation of turn-level performances}.
\newblock In \emph{Proceedings of the 60th Annual Meeting of the Association for Computational Linguistics (Volume 2: Short Papers)}, pages 318--324, Dublin, Ireland. Association for Computational Linguistics.

\bibitem[{Doshi-Velez et~al.(2019)Doshi-Velez, Kortz, Budish, Bavitz, Gershman, O'Brien, Scott, Schieber, Waldo, Weinberger, Weller, and Wood}]{accountability}
Finale Doshi-Velez, Mason Kortz, Ryan Budish, Chris Bavitz, Sam Gershman, David O'Brien, Kate Scott, Stuart Schieber, James Waldo, David Weinberger, Adrian Weller, and Alexandra Wood. 2019.
\newblock \href {https://arxiv.org/abs/1711.01134} {Accountability of ai under the law: The role of explanation}.
\newblock \emph{Preprint}, arXiv:1711.01134.

\bibitem[{Dubey and et~al.(2024)}]{llama}
Abhimanyu Dubey and et~al. 2024.
\newblock \href {https://arxiv.org/abs/2407.21783} {The llama 3 herd of models}.
\newblock \emph{Preprint}, arXiv:2407.21783.

\bibitem[{Feng et~al.(2023{\natexlab{a}})Feng, Lu, Liu, Zhan, and Wu}]{llm-dst}
Yujie Feng, Zexin Lu, Bo~Liu, Liming Zhan, and Xiao-Ming Wu. 2023{\natexlab{a}}.
\newblock \href {https://doi.org/10.18653/v1/2023.emnlp-main.48} {Towards {LLM}-driven dialogue state tracking}.
\newblock In \emph{Proceedings of the 2023 Conference on Empirical Methods in Natural Language Processing}, pages 739--755, Singapore. Association for Computational Linguistics.

\bibitem[{Feng et~al.(2023{\natexlab{b}})Feng, Lu, Liu, Zhan, and Wu}]{ldst}
Yujie Feng, Zexin Lu, Bo~Liu, Liming Zhan, and Xiao-Ming Wu. 2023{\natexlab{b}}.
\newblock \href {https://doi.org/10.18653/v1/2023.emnlp-main.48} {Towards {LLM}-driven dialogue state tracking}.
\newblock In \emph{Proceedings of the 2023 Conference on Empirical Methods in Natural Language Processing}, pages 739--755, Singapore. Association for Computational Linguistics.

\bibitem[{Gao et~al.(2019)Gao, Sethi, Agarwal, Chung, and Hakkani-Tur}]{gao-reading}
Shuyang Gao, Abhishek Sethi, Sanchit Agarwal, Tagyoung Chung, and Dilek Hakkani-Tur. 2019.
\newblock \href {https://doi.org/10.18653/v1/W19-5932} {Dialog state tracking: A neural reading comprehension approach}.
\newblock In \emph{Proceedings of the 20th Annual SIGdial Meeting on Discourse and Dialogue}, pages 264--273, Stockholm, Sweden. Association for Computational Linguistics.

\bibitem[{Heck et~al.(2022)Heck, Lubis, van Niekerk, Feng, Geishauser, Lin, and Ga{\v{s}}i{\'c}}]{heck-etal-2022-robust}
Michael Heck, Nurul Lubis, Carel van Niekerk, Shutong Feng, Christian Geishauser, Hsien-Chin Lin, and Milica Ga{\v{s}}i{\'c}. 2022.
\newblock \href {https://doi.org/10.1162/tacl_a_00513} {Robust dialogue state tracking with weak supervision and sparse data}.
\newblock \emph{Transactions of the Association for Computational Linguistics}, 10:1175--1192.

\bibitem[{Heck et~al.(2020)Heck, van Niekerk, Lubis, Geishauser, Lin, Moresi, and Gasic}]{trippy}
Michael Heck, Carel van Niekerk, Nurul Lubis, Christian Geishauser, Hsien-Chin Lin, Marco Moresi, and Milica Gasic. 2020.
\newblock \href {https://www.aclweb.org/anthology/2020.sigdial-1.4} {{T}rip{P}y: A triple copy strategy for value independent neural dialog state tracking}.
\newblock In \emph{Proceedings of the 21th Annual Meeting of the Special Interest Group on Discourse and Dialogue}, pages 35--44, 1st virtual meeting. Association for Computational Linguistics.

\bibitem[{Henderson et~al.(2014)Henderson, Thomson, and Williams}]{dstc2}
Matthew Henderson, Blaise Thomson, and Jason~D. Williams. 2014.
\newblock \href {https://doi.org/10.3115/v1/W14-4337} {The second dialog state tracking challenge}.
\newblock In \emph{Proceedings of the 15th Annual Meeting of the Special Interest Group on Discourse and Dialogue ({SIGDIAL})}, pages 263--272, Philadelphia, PA, U.S.A. Association for Computational Linguistics.

\bibitem[{Hosseini-Asl et~al.(2020)Hosseini-Asl, McCann, Wu, Yavuz, and Socher}]{simpletod}
Ehsan Hosseini-Asl, Bryan McCann, Chien-Sheng Wu, Semih Yavuz, and Richard Socher. 2020.
\newblock \href {https://proceedings.neurips.cc/paper/2020/file/e946209592563be0f01c844ab2170f0c-Paper.pdf} {A simple language model for task-oriented dialogue}.
\newblock In \emph{Advances in Neural Information Processing Systems}, volume~33, pages 20179--20191. Curran Associates, Inc.

\bibitem[{Hu et~al.(2021)Hu, Shen, Wallis, Allen-Zhu, Li, Wang, Wang, and Chen}]{lora}
Edward~J. Hu, Yelong Shen, Phillip Wallis, Zeyuan Allen-Zhu, Yuanzhi Li, Shean Wang, Lu~Wang, and Weizhu Chen. 2021.
\newblock \href {https://arxiv.org/abs/2106.09685} {Lora: Low-rank adaptation of large language models}.
\newblock \emph{Preprint}, arXiv:2106.09685.

\bibitem[{Hu et~al.(2022)Hu, Lee, Xie, Yu, Smith, and Ostendorf}]{ic-dst}
Yushi Hu, Chia-Hsuan Lee, Tianbao Xie, Tao Yu, Noah~A. Smith, and Mari Ostendorf. 2022.
\newblock \href {https://doi.org/10.18653/v1/2022.findings-emnlp.193} {In-context learning for few-shot dialogue state tracking}.
\newblock In \emph{Findings of the Association for Computational Linguistics: EMNLP 2022}, pages 2627--2643, Abu Dhabi, United Arab Emirates. Association for Computational Linguistics.

\bibitem[{Huang et~al.(2024)Huang, Chen, Mishra, Zheng, Yu, Song, and Zhou}]{huang2024large}
Jie Huang, Xinyun Chen, Swaroop Mishra, Huaixiu~Steven Zheng, Adams~Wei Yu, Xinying Song, and Denny Zhou. 2024.
\newblock \href {https://openreview.net/forum?id=IkmD3fKBPQ} {Large language models cannot self-correct reasoning yet}.
\newblock In \emph{The Twelfth International Conference on Learning Representations}.

\bibitem[{Huang et~al.(2023)Huang, Yu, Ma, Zhong, Feng, Wang, Chen, Peng, Feng, Qin, and Liu}]{llm-hal}
Lei Huang, Weijiang Yu, Weitao Ma, Weihong Zhong, Zhangyin Feng, Haotian Wang, Qianglong Chen, Weihua Peng, Xiaocheng Feng, Bing Qin, and Ting Liu. 2023.
\newblock \href {https://arxiv.org/abs/2311.05232} {A survey on hallucination in large language models: Principles, taxonomy, challenges, and open questions}.
\newblock \emph{Preprint}, arXiv:2311.05232.

\bibitem[{Hude{\v{c}}ek and Dusek(2023)}]{tods-llm}
Vojt{\v{e}}ch Hude{\v{c}}ek and Ondrej Dusek. 2023.
\newblock \href {https://doi.org/10.18653/v1/2023.sigdial-1.21} {Are large language models all you need for task-oriented dialogue?}
\newblock In \emph{Proceedings of the 24th Annual Meeting of the Special Interest Group on Discourse and Dialogue}, pages 216--228, Prague, Czechia. Association for Computational Linguistics.

\bibitem[{Ji et~al.(2023)Ji, Lee, Frieske, Yu, Su, Xu, Ishii, Bang, Madotto, and Fung}]{nlg-hal}
Ziwei Ji, Nayeon Lee, Rita Frieske, Tiezheng Yu, Dan Su, Yan Xu, Etsuko Ishii, Ye~Jin Bang, Andrea Madotto, and Pascale Fung. 2023.
\newblock \href {https://doi.org/10.1145/3571730} {Survey of hallucination in natural language generation}.
\newblock \emph{ACM Comput. Surv.}, 55(12).

\bibitem[{Jiang et~al.(2023)Jiang, Sablayrolles, Mensch, Bamford, Chaplot, de~las Casas, Bressand, Lengyel, Lample, Saulnier, Lavaud, Lachaux, Stock, Scao, Lavril, Wang, Lacroix, and Sayed}]{mistral}
Albert~Q. Jiang, Alexandre Sablayrolles, Arthur Mensch, Chris Bamford, Devendra~Singh Chaplot, Diego de~las Casas, Florian Bressand, Gianna Lengyel, Guillaume Lample, Lucile Saulnier, Lélio~Renard Lavaud, Marie-Anne Lachaux, Pierre Stock, Teven~Le Scao, Thibaut Lavril, Thomas Wang, Timothée Lacroix, and William~El Sayed. 2023.
\newblock \href {https://arxiv.org/abs/2310.06825} {Mistral 7b}.
\newblock \emph{Preprint}, arXiv:2310.06825.

\bibitem[{Kim et~al.(2020)Kim, Yang, Kim, and Lee}]{som-dst}
Sungdong Kim, Sohee Yang, Gyuwan Kim, and Sang-Woo Lee. 2020.
\newblock \href {https://doi.org/10.18653/v1/2020.acl-main.53} {Efficient dialogue state tracking by selectively overwriting memory}.
\newblock In \emph{Proceedings of the 58th Annual Meeting of the Association for Computational Linguistics}, pages 567--582, Online. Association for Computational Linguistics.

\bibitem[{Klingbeil et~al.(2024)Klingbeil, Grützner, and Schreck}]{trust-reliance-ai}
Artur Klingbeil, Cassandra Grützner, and Philipp Schreck. 2024.
\newblock \href {https://doi.org/10.1016/j.chb.2024.108352} {Trust and reliance on ai — an experimental study on the extent and costs of overreliance on ai}.
\newblock \emph{Computers in Human Behavior}, 160:108352.

\bibitem[{Lee et~al.(2024{\natexlab{a}})Lee, Cheng, and Ostendorf}]{correctionlm}
Chia-Hsuan Lee, Hao Cheng, and Mari Ostendorf. 2024{\natexlab{a}}.
\newblock \href {https://arxiv.org/abs/2410.18209} {Correctionlm: Self-corrections with slm for dialogue state tracking}.
\newblock \emph{Preprint}, arXiv:2410.18209.

\bibitem[{Lee et~al.(2024{\natexlab{b}})Lee, Cheng, and Ostendorf}]{orchestrallm}
Chia-Hsuan Lee, Hao Cheng, and Mari Ostendorf. 2024{\natexlab{b}}.
\newblock \href {https://doi.org/10.18653/v1/2024.naacl-long.79} {{O}rchestra{LLM}: Efficient orchestration of language models for dialogue state tracking}.
\newblock In \emph{Proceedings of the 2024 Conference of the North American Chapter of the Association for Computational Linguistics: Human Language Technologies (Volume 1: Long Papers)}, pages 1434--1445, Mexico City, Mexico. Association for Computational Linguistics.

\bibitem[{Lee et~al.(2019)Lee, Lee, and Kim}]{sumbt}
Hwaran Lee, Jinsik Lee, and Tae-Yoon Kim. 2019.
\newblock \href {https://doi.org/10.18653/v1/P19-1546} {{SUMBT}: Slot-utterance matching for universal and scalable belief tracking}.
\newblock In \emph{Proceedings of the 57th Annual Meeting of the Association for Computational Linguistics}, pages 5478--5483, Florence, Italy. Association for Computational Linguistics.

\bibitem[{Li et~al.(2024{\natexlab{a}})Li, Wang, Feng, Zhu, Wang, and Chua}]{thinktwice}
Moxin Li, Wenjie Wang, Fuli Feng, Fengbin Zhu, Qifan Wang, and Tat-Seng Chua. 2024{\natexlab{a}}.
\newblock \href {https://arxiv.org/abs/2403.09972} {Think twice before trusting: Self-detection for large language models through comprehensive answer reflection}.
\newblock \emph{Preprint}, arXiv:2403.09972.

\bibitem[{Li et~al.(2024{\natexlab{b}})Li, Chen, Ross, Huber, Moon, Lin, Dong, Sagar, Yan, and Crook}]{fnctod}
Zekun Li, Zhiyu Chen, Mike Ross, Patrick Huber, Seungwhan Moon, Zhaojiang Lin, Xin Dong, Adithya Sagar, Xifeng Yan, and Paul Crook. 2024{\natexlab{b}}.
\newblock \href {https://doi.org/10.18653/v1/2024.acl-long.471} {Large language models as zero-shot dialogue state tracker through function calling}.
\newblock In \emph{Proceedings of the 62nd Annual Meeting of the Association for Computational Linguistics (Volume 1: Long Papers)}, pages 8688--8704, Bangkok, Thailand. Association for Computational Linguistics.

\bibitem[{Lin et~al.(2020)Lin, Madotto, Winata, and Fung}]{mintl}
Zhaojiang Lin, Andrea Madotto, Genta~Indra Winata, and Pascale Fung. 2020.
\newblock \href {https://doi.org/10.18653/v1/2020.emnlp-main.273} {{M}in{TL}: Minimalist transfer learning for task-oriented dialogue systems}.
\newblock In \emph{Proceedings of the 2020 Conference on Empirical Methods in Natural Language Processing (EMNLP)}, pages 3391--3405, Online. Association for Computational Linguistics.

\bibitem[{Mehri et~al.(2020)Mehri, Eric, and Hakkani-Tur}]{dialoglue}
Shikib Mehri, Mihail Eric, and Dilek Hakkani-Tur. 2020.
\newblock \href {https://arxiv.org/abs/2009.13570} {Dialoglue: A natural language understanding benchmark for task-oriented dialogue}.
\newblock \emph{Preprint}, arXiv:2009.13570.

\bibitem[{Mejtoft et~al.(2019)Mejtoft, Hale, and S\"{o}derstr\"{o}m}]{design-friction}
Thomas Mejtoft, Sarah Hale, and Ulrik S\"{o}derstr\"{o}m. 2019.
\newblock \href {https://doi.org/10.1145/3335082.3335106} {Design friction}.
\newblock ECCE '19, page 41–44, New York, NY, USA. Association for Computing Machinery.

\bibitem[{Mrk{\v{s}}i{\'c} et~al.(2017)Mrk{\v{s}}i{\'c}, {\'O}~S{\'e}aghdha, Wen, Thomson, and Young}]{nbt}
Nikola Mrk{\v{s}}i{\'c}, Diarmuid {\'O}~S{\'e}aghdha, Tsung-Hsien Wen, Blaise Thomson, and Steve Young. 2017.
\newblock \href {https://doi.org/10.18653/v1/P17-1163} {Neural belief tracker: Data-driven dialogue state tracking}.
\newblock In \emph{Proceedings of the 55th Annual Meeting of the Association for Computational Linguistics (Volume 1: Long Papers)}, pages 1777--1788, Vancouver, Canada. Association for Computational Linguistics.

\bibitem[{Naiseh et~al.(2021)Naiseh, Al-Mansoori, Al-Thani, Jiang, and Ali}]{nudge-friction}
Mohammad Naiseh, Reem~S. Al-Mansoori, Dena Al-Thani, Nan Jiang, and Raian Ali. 2021.
\newblock \href {https://doi.org/10.1109/BESC53957.2021.9635271} {Nudging through friction: An approach for calibrating trust in explainable ai}.
\newblock In \emph{2021 8th International Conference on Behavioral and Social Computing (BESC)}, pages 1--5.

\bibitem[{Nouri and Hosseini-Asl(2018)}]{gce}
Elnaz Nouri and Ehsan Hosseini-Asl. 2018.
\newblock \href {https://arxiv.org/abs/1812.00899} {Toward scalable neural dialogue state tracking}.
\newblock In \emph{NeurIPS 2018, 2nd Conversational AI workshop}.

\bibitem[{Novelli et~al.(2022)Novelli, Taddeo, and Floridi}]{account-ai}
Claudio Novelli, Mariarosaria Taddeo, and Luciano Floridi. 2022.
\newblock \href {https://doi.org/10.1007/s00146-023-01635-y} {Accountability in artificial intelligence: What it is and how it works}.
\newblock \emph{AI \& Society: Journal of Knowledge, Culture and Communication - Springer}.

\bibitem[{OpenAI and et~al.(2024)}]{openai2024gpt4technicalreport}
OpenAI and et~al. 2024.
\newblock \href {https://arxiv.org/abs/2303.08774} {Gpt-4 technical report}.
\newblock \emph{Preprint}, arXiv:2303.08774.

\bibitem[{Passi and Vorvoreanu(2022)}]{ai-overreliance}
Samir Passi and Mihaela Vorvoreanu. 2022.
\newblock Overreliance on ai: Literature review.
\newblock Technical report, Microsoft Technical Report MSR-TR-2022-12. Microsoft Corporation.

\bibitem[{Su et~al.(2022)Su, Shu, Mansimov, Gupta, Cai, Lai, and Zhang}]{pptod}
Yixuan Su, Lei Shu, Elman Mansimov, Arshit Gupta, Deng Cai, Yi-An Lai, and Yi~Zhang. 2022.
\newblock \href {https://doi.org/10.18653/v1/2022.acl-long.319} {Multi-task pre-training for plug-and-play task-oriented dialogue system}.
\newblock In \emph{Proceedings of the 60th Annual Meeting of the Association for Computational Linguistics (Volume 1: Long Papers)}, pages 4661--4676, Dublin, Ireland. Association for Computational Linguistics.

\bibitem[{Sun et~al.(2024)Sun, Dey, Hakkani-Tur, and Tur}]{conf-dst}
Yi-Jyun Sun, Suvodip Dey, Dilek Hakkani-Tur, and Gokhan Tur. 2024.
\newblock \href {https://arxiv.org/abs/2409.09629} {Confidence estimation for llm-based dialogue state tracking}.
\newblock \emph{Preprint}, arXiv:2409.09629.

\bibitem[{Team and et~al.(2024)}]{gemma}
Gemma Team and et~al. 2024.
\newblock \href {https://arxiv.org/abs/2403.08295} {Gemma: Open models based on gemini research and technology}.
\newblock \emph{Preprint}, arXiv:2403.08295.

\bibitem[{Vaswani et~al.(2017)Vaswani, Shazeer, Parmar, Uszkoreit, Jones, Gomez, Kaiser, and Polosukhin}]{transformer}
Ashish Vaswani, Noam Shazeer, Niki Parmar, Jakob Uszkoreit, Llion Jones, Aidan~N. Gomez, undefinedukasz Kaiser, and Illia Polosukhin. 2017.
\newblock \href {https://papers.nips.cc/paper/2017/file/3f5ee243547dee91fbd053c1c4a845aa-Paper.pdf} {Attention is all you need}.
\newblock In \emph{Proceedings of the 31st International Conference on Neural Information Processing Systems}, NIPS'17, page 6000–6010, Red Hook, NY, USA. Curran Associates Inc.

\bibitem[{Wu et~al.(2019)Wu, Madotto, Hosseini-Asl, Xiong, Socher, and Fung}]{trade}
Chien-Sheng Wu, Andrea Madotto, Ehsan Hosseini-Asl, Caiming Xiong, Richard Socher, and Pascale Fung. 2019.
\newblock \href {https://doi.org/10.18653/v1/P19-1078} {Transferable multi-domain state generator for task-oriented dialogue systems}.
\newblock In \emph{Proceedings of the 57th Annual Meeting of the Association for Computational Linguistics}, pages 808--819, Florence, Italy. Association for Computational Linguistics.

\bibitem[{Xiong et~al.(2024)Xiong, Hu, Lu, LI, Fu, He, and Hooi}]{xiong2024can}
Miao Xiong, Zhiyuan Hu, Xinyang Lu, YIFEI LI, Jie Fu, Junxian He, and Bryan Hooi. 2024.
\newblock \href {https://openreview.net/forum?id=gjeQKFxFpZ} {Can {LLM}s express their uncertainty? an empirical evaluation of confidence elicitation in {LLM}s}.
\newblock In \emph{The Twelfth International Conference on Learning Representations}.

\bibitem[{Xu et~al.(2024)Xu, Mao, Yang, Sun, and Huang}]{autotod}
Heng-Da Xu, Xian-Ling Mao, Puhai Yang, Fanshu Sun, and Heyan Huang. 2024.
\newblock \href {https://doi.org/10.18653/v1/2024.acl-long.152} {Rethinking task-oriented dialogue systems: From complex modularity to zero-shot autonomous agent}.
\newblock In \emph{Proceedings of the 62nd Annual Meeting of the Association for Computational Linguistics (Volume 1: Long Papers)}, pages 2748--2763, Bangkok, Thailand. Association for Computational Linguistics.

\bibitem[{Ye et~al.(2022{\natexlab{a}})Ye, Feng, and Yilmaz}]{assist}
Fanghua Ye, Yue Feng, and Emine Yilmaz. 2022{\natexlab{a}}.
\newblock \href {https://doi.org/10.18653/v1/2022.findings-acl.214} {{ASSIST}: Towards label noise-robust dialogue state tracking}.
\newblock In \emph{Findings of the Association for Computational Linguistics: ACL 2022}, pages 2719--2731, Dublin, Ireland. Association for Computational Linguistics.

\bibitem[{Ye et~al.(2022{\natexlab{b}})Ye, Manotumruksa, and Yilmaz}]{multiwoz24}
Fanghua Ye, Jarana Manotumruksa, and Emine Yilmaz. 2022{\natexlab{b}}.
\newblock \href {https://doi.org/10.18653/v1/2022.sigdial-1.34} {{M}ulti{WOZ} 2.4: A multi-domain task-oriented dialogue dataset with essential annotation corrections to improve state tracking evaluation}.
\newblock In \emph{Proceedings of the 23rd Annual Meeting of the Special Interest Group on Discourse and Dialogue}, pages 351--360, Edinburgh, UK. Association for Computational Linguistics.

\bibitem[{Ye et~al.(2021)Ye, Manotumruksa, Zhang, Li, and Yilmaz}]{star}
Fanghua Ye, Jarana Manotumruksa, Qiang Zhang, Shenghui Li, and Emine Yilmaz. 2021.
\newblock \href {https://arxiv.org/abs/2101.09374} {Slot self-attentive dialogue state tracking}.
\newblock \emph{Preprint}, arXiv:2101.09374.

\bibitem[{Ye et~al.(2022{\natexlab{c}})Ye, Wang, Huang, Li, Stern, and Yilmaz}]{metaassist}
Fanghua Ye, Xi~Wang, Jie Huang, Shenghui Li, Samuel Stern, and Emine Yilmaz. 2022{\natexlab{c}}.
\newblock \href {https://doi.org/10.18653/v1/2022.emnlp-main.76} {{M}eta{ASSIST}: Robust dialogue state tracking with meta learning}.
\newblock In \emph{Proceedings of the 2022 Conference on Empirical Methods in Natural Language Processing}, pages 1157--1169, Abu Dhabi, United Arab Emirates. Association for Computational Linguistics.

\bibitem[{Zhao et~al.(2021)Zhao, Mahdieh, Zhang, Cao, and Wu}]{zhao-etal-2021-effective-sequence}
Jeffrey Zhao, Mahdis Mahdieh, Ye~Zhang, Yuan Cao, and Yonghui Wu. 2021.
\newblock \href {https://doi.org/10.18653/v1/2021.emnlp-main.593} {Effective sequence-to-sequence dialogue state tracking}.
\newblock In \emph{Proceedings of the 2021 Conference on Empirical Methods in Natural Language Processing}, pages 7486--7493, Online and Punta Cana, Dominican Republic. Association for Computational Linguistics.

\bibitem[{Zhong et~al.(2018)Zhong, Xiong, and Socher}]{glad}
Victor Zhong, Caiming Xiong, and Richard Socher. 2018.
\newblock \href {https://doi.org/10.18653/v1/P18-1135} {Global-locally self-attentive encoder for dialogue state tracking}.
\newblock In \emph{Proceedings of the 56th Annual Meeting of the Association for Computational Linguistics (Volume 1: Long Papers)}, pages 1458--1467, Melbourne, Australia. Association for Computational Linguistics.

\bibitem[{İnan et~al.(2025)İnan, Sicilia, Dey, Dongre, Srinivasan, Thomason, Tür, Hakkani-Tür, and Alikhani}]{i̇nan2025betterslowsorryintroducing}
Mert İnan, Anthony Sicilia, Suvodip Dey, Vardhan Dongre, Tejas Srinivasan, Jesse Thomason, Gökhan Tür, Dilek Hakkani-Tür, and Malihe Alikhani. 2025.
\newblock \href {https://arxiv.org/abs/2501.17348} {Better slow than sorry: Introducing positive friction for reliable dialogue systems}.
\newblock \emph{Preprint}, arXiv:2501.17348.

\end{thebibliography}

\appendix

\section{Appendix}

\begin{table}[b]
\centering
\begin{tabular}{r|r|r}
\multirow{2}{*}{\textbf{$\lambda$}} & \multicolumn{2}{c}{\textbf{JGA}}
\\ \cline{2-3}
\centering 
 & \textbf{MultiWOZ} & \textbf{Snips} \\ \hline
0 & 64.24 & 91.14 \\ 
0.1 & 66.13 & 92.00 \\ 
0.25 & \textbf{66.57} & \textbf{92.86}  \\ 
0.5 & 66.25 & 92.43  \\ 
0.75 & 65.34 & 92.04  \\ 
1.0 & 65.15 & 91.76  \\ 
\hline

\end{tabular}
\caption{Impact of varying $\lambda$ in $\mathcal{M_{\operatorname{account}}}$ with Llama backbone on validation data.}
\label{tbl:hyp1}
\vspace{-0.1in}
\end{table}

\subsection{Selection of $\lambda$}
\label{sec:hyp}
 Table~\ref{tbl:hyp1} shows the joint goal accuracy for different $\lambda$ (in Eqn.~\ref{eqn:obj}) on both the MultiWOZ and Snips validation sets. Table~\ref{tbl:hyp1} shows the results with the Llama 3.1 model. We can observe that $\lambda=0.25$ results in the best validation performance in our setup. It also results in the best validation performance for Mistral and Gemma on the MultiWOZ dataset. However, Mistral and Gemma show the best validation performance for the Snips dataset with $\lambda=0.1$.

\subsection{Additional Details}
\label{sec:prompt}
This section provides additional details related to model training, inference, and prompts. The experiments are performed on Nvidia A100 machines. The hidden size of Llama and Mistral is 4096 and 3072 for Gemma. The number of trainable parameters in our LoRA fine-tuned models is 3.6M for Llama and Mistral. On the other hand, the trainable parameters of the Gemma model is 3.4M. We found that 8B models provided an optimal balance for training on our GPU servers. Smaller models resulted in lower performance, while larger models posed computational challenges due to resource constraints. This is why we conducted our experiments using LLaMA 3.1 (8B), Mistral (7B), and Gemma (7B). It also helped to balance between performance and computational efficiency, making our approach both scalable and practical for real-world applications. Due to memory limitations in GPU, we use a batch size of 1. We increase the effective batch size to 8 by using gradient accumulation. The training time for MultiWOZ and Snips is approximately 20 and 6 GPU hours, respectively. 

During inference, the model takes only the dialogue context as input and generates the corresponding dialogue state, without relying on any oracle belief state. We do not apply any additional post-processing to enforce JSON formatting. In our experiments, all outputs adhered to the expected JSON format, and no formatting issues were observed. However, in the event of a failure, the output will default to an empty dictionary (i.e., \{\}). All the belief states are generated using a single run. 

Table~\ref{tbl:prompt1} shows the prompt format used to formulate the dialogue context ($C_t$) to generate the belief states. It contains the task information, slot descriptions, dialogue history, and output format. Note that Table~\ref{tbl:prompt1} shows the prompt specific to Llama. The prompt remains exactly the same for Mistral and Gemma, except for the special tokens. For the Snips dataset, we use the same template with the Snips-specific slot description.

In Section~\ref{sec:overreliance}, we used two kinds of prompts to confirm the predicted errors from the user-simulator. The prompt for confirming false positives is shown in Table~\ref{tbl:prompt2}, while the prompt for confirming false negatives is shown in Table~\ref{tbl:prompt3}.

\begin{table*}[t]
\begin{small}
\centering
\begin{tabular}{|m{15cm}|}
\hline
<|begin\_of\_text|>You are a helpful assistant who can perform dialogue-state tracking. The user interacts with the system to book entities from multiple domains (hotel, restaurant, attraction, taxi, and train) in Cambridge. Your goal is to find all the intents shown by the user in the conversation.\\ \\

The user can ask for a hotel by slots - hotel-name, hotel-type, hotel-parking, hotel-area, hotel-bookday, hotel-bookstay, hotel-internet, hotel-bookpeople, hotel-stars, hotel-pricerange. The user can ask for an attraction by slots - attraction-name, attraction-type, attraction-area. The user can ask for a restaurant by slots - restaurant-name, restaurant-food, restaurant-area, restaurant-bookday, restaurant-booktime, restaurant-bookpeople, restaurant-pricerange. The user can ask for a taxi by slots - taxi-arriveby, taxi-departure, taxi-leaveat, taxi-destination. The user can ask for a train by slots - train-arriveby, train-day, train-leaveat, train-destination, train-departure, train-bookpeople. Do not capture any other slots!\\ \\

\# Task\\
You will be provided with a chronological dialogue history between the system and the user. You must find all the user intents and output them in JSON format.\\ \\

\# Sample Output \\
\{"restaurant-name": "abc", "restaurant-food": "xyz"\}\\

\# Conversation History \\
<|start\_header\_id|>system<|end\_header\_id|>\\
$S_0$<|eot\_id|>\\
<|start\_header\_id|>user<|end\_header\_id|>\\
$U_0$<|eot\_id|>\\
...\\

<|start\_header\_id|>assistant<|end\_header\_id|>\\

\hline
\end{tabular}
\caption{Prompt template for Llama model. $S_0$ and $U_0$ denote the system and user utterance for the Turn $0$.}
\label{tbl:prompt1}
\end{small}
\end{table*} 

\begin{table*}[ht!]
\begin{small}
\centering
\begin{tabular}{|m{15cm}|}
\hline
Dialogue history:\\
\{Current Dialogue History\}\\ \\

It seems that some information might have been incorrectly predicted.
Based on the current user utterance, please help us clarify the following:\\ \\

1. If the slot name is correct but the value is wrong, you can provide the correct value.\\
2. If the slot name should not appear at all, let us know to delete it.\\ \\

\# Statement: {Slot} was mentioned in the dialogue history and the value is {Value}.\\

- Please respond with one of the following options:\\
  - "Agree" (if both the slot name and value are correct)\\
  - "Not Agree: Update to [new value]" (if the slot name is correct but the value is wrong)\\
  - "Not Agree: Delete" (if the slot name should not appear in the belief state)\\
\hline
\end{tabular}
\caption{Prompt template for clarifying false positive with user simulator.}
\label{tbl:prompt2}
\end{small}
\vspace{-0.1in}
\end{table*} 



\begin{table*}[ht!]
\begin{small}
\centering
\begin{tabular}{|m{15cm}|}
\hline
Dialogue history:\\
{Current Dialogue History}\\ \\

It seems that some important information might be missing. \\
Based on the latest user utterance, could you please confirm the following:\\

Should the {Slot} be "{Value}"?\\ \\

- Please respond with "Agree" if this information is correct. \\
- Respond with "Not Agree" if this information is incorrect.\\ \\

Your confirmation will help us improve the accuracy of the prediction. \\
\hline
\end{tabular}
\caption{Prompt template for clarifying false negative with user simulator.}
\label{tbl:prompt3}
\end{small}
\vspace{-0.1in}
\end{table*}

\end{document}